\documentclass{article}

\usepackage[utf8]{inputenc} 
\usepackage[T1]{fontenc}    
\usepackage{booktabs}       
\usepackage{nicefrac}       
\usepackage{microtype}      
\usepackage{times}             

\usepackage[round]{natbib}
\usepackage{amssymb,amsmath,amsthm,bbm}
\usepackage[margin=1in]{geometry}
\usepackage{verbatim,float,url,dsfont}
\usepackage{graphicx,subfigure,psfrag}
\usepackage{algorithm,algorithmic}
\usepackage{mathtools,enumitem}
\usepackage[colorlinks,citecolor=blue]{hyperref}
\usepackage[utf8]{inputenc} 
\usepackage[T1]{fontenc}    
\usepackage{hyperref}       
\usepackage{url}            
\usepackage{booktabs}       
\usepackage{amsfonts}       
\usepackage{nicefrac}       
\usepackage{microtype}      

\usepackage{times}             
\usepackage{amsthm}

\usepackage{bbm}
\usepackage{verbatim,float,dsfont}
\usepackage{psfrag}
\usepackage{algorithm,algorithmic}
\usepackage{mathtools,enumitem}
\mathtoolsset{showonlyrefs}
\usepackage{tcolorbox}
\usepackage{breqn,lipsum}
\usepackage{thmtools,thm-restate}
\usepackage{graphicx,stackengine}
\usepackage{array}
\usepackage{caption}
\usepackage{boldline}
\usepackage{amssymb}
\usepackage{multirow}

\newtheorem{theorem}{Theorem}
\newtheorem{lemma}[theorem]{Lemma}
\newtheorem{corollary}[theorem]{Corollary}
\newtheorem{proposition}[theorem]{Proposition}
\newtheorem{remark}[theorem]{Remark}
\newtheorem{definition}[theorem]{Definition}

\DeclarePairedDelimiter\floor{\lfloor}{\rfloor}

\def\shownotes{1}  
\ifnum\shownotes=1
\newcommand{\authnote}[2]{$\ll$\textsf{\footnotesize #1 notes: #2}$\gg$}
\else
\newcommand{\authnote}[2]{}
\fi

\usepackage{accents}

\makeatletter
\newcommand*\rel@kern[1]{\kern#1\dimexpr\macc@kerna}
\newcommand*\widebar[1]{%
  \begingroup
  \def\mathaccent##1##2{%
    \rel@kern{0.8}%
    \overline{\rel@kern{-0.8}\macc@nucleus\rel@kern{0.2}}%
    \rel@kern{-0.2}%
  }%
  \macc@depth\@ne
  \let\math@bgroup\@empty \let\math@egroup\macc@set@skewchar
  \mathsurround\z@ \frozen@everymath{\mathgroup\macc@group\relax}%
  \macc@set@skewchar\relax
  \let\mathaccentV\macc@nested@a
  \macc@nested@a\relax111{#1}%
  \endgroup
}
\makeatother

\newcommand{\argmin}{\mathop{\mathrm{argmin}}}
\newcommand{\argmax}{\mathop{\mathrm{argmax}}}

\def\E{\mathbb{E}}

\def\row{\mathrm{row}}

\def\cA{\mathcal{A}}

\def\cD{\mathcal{D}}
\def\cE{\mathcal{E}}
\def\cF{\mathcal{F}}

\def\cM{\mathcal{M}}

\def\TV{\mathrm{TV}}

\makeatletter

\newcommand*\totht[1]{\dimexpr\ht#1+\dp#1\relax}
\newcommand*\leading{{\setbox0\hbox{\strut}\the\totht0}}
\newcommand*\fntsize{{\setbox0\hbox{Mg}\the\totht0}}
\newcommand*\showsize[1]{{#1 {\ttfamily\string#1} (\f@size pt) \fntsize/\leading}\par}
\makeatother

\newcommand{\grad}{\nabla}
\newcommand{\lamda}{\lambda}

\title{Optimal Dynamic Regret in LQR Control}

\author{Dheeraj Baby \\dheeraj@ucsb.edu \and Yu-Xiang Wang \\yuxiangw@cs.ucsb.edu}

\date{Dept. of Computer Science \\ UC Santa Barbara}

\begin{document}

\maketitle

\begin{abstract}
We consider the problem of nonstochastic control with a sequence of quadratic losses, i.e., LQR control. We provide an efficient online algorithm that achieves an optimal dynamic (policy) regret of $\tilde{O}(\max\{n^{1/3} \TV(M_{1:n})^{2/3}, 1\})$, where $\TV(M_{1:n})$ is the total variation of any oracle sequence of \emph{Disturbance Action} policies parameterized by $M_1,...,M_n$ --- chosen in hindsight to cater to unknown nonstationarity. The rate improves the best known rate of $\tilde{O}(\sqrt{n (\TV(M_{1:n})+1)} )$ for general convex losses and we prove that it is information-theoretically optimal for LQR. Main technical components include the reduction of LQR to online linear regression with delayed feedback due to \citet{Foster2020LogarithmicLQR}, as well as a new \emph{proper} learning algorithm with an optimal $\tilde{O}(n^{1/3})$ dynamic regret on a family of ``minibatched'' quadratic losses, which could be of independent interest.
\end{abstract}

\section{Introduction}
This paper studies the linear quadratic regulator (LQR) control problem which is a specific instantiation of the more general RL framework where the evolution of states follows a predefined linear dynamics. At each round $t \in [n]:=\{1,\ldots,n \}$, the agent is at state $x_t \in \mathbb{R}^{d_x}$. Based on the state, the agent select a control input $u_t \in \mathbb{R}^{d_u}$. The next state evolves according to the law:
\begin{align}
    x_{t+1} = Ax_t + Bu_t + w_t, \label{eq:state-evolv}
\end{align}

where $A$ and $B$ are system matrices known to the agent. $w_t \in \mathbb{R}^{d_x}$ is a disturbance term that can be selected by a potentially adaptive adversary. We assume that $\| w_t\|_2 \le 1$. This disturbance term reflects the perturbation from the ideal linear state transition arising due to environmental factors that could be difficult to model. The loss suffered by playing the control $u$ at state $x$ is given by $\ell(x,u) := x^TR_xx + u^TR_uu$, where $R_x,R_u \succcurlyeq 0$, that are apriori fixed and known.

Recently there has been a surge of interest in viewing this classical LQR problem under the lens of online learning \citep{hazan2016introduction}. The work of \citet{agarwal2019control} places regret of the agent against a set of benchmark policies as the central notion to evaluate learner's performance. Following \citet{agarwal2019control,Foster2020LogarithmicLQR} we adopt the  class of disturbance action policies (DAP) as our benchmark class:
\begin{definition} \label{def:dap} (Disturbance action policies, \citep{Foster2020LogarithmicLQR}). Let $M=(M^{[i]})_{i=1}^m$ denote a sequence of matrices $M^{[i]} \in \mathbb{R}^{d_u \times d_x}$. We define the corresponding disturbance action policies (DAP) $\pi^{M}$ as:
\begin{align}
    \pi_t^{M}(x_t)
    &= -K_\infty x_t - q^M(w_{1:t-1}), \label{eq:policy} 
\end{align}
where $q^M(w_{1:t-1}) = \sum_{i=1}^m M^{[i]} w_{t-i}$ and $K_\infty$ as in Eq.\eqref{eq:kinf}. We are interested in DAPs for which the sequence $M$ belongs to the set:
\begin{align}
    \mathcal{M}(m,R,\gamma) := \{M=(M^{[i]})_{i=1}^m : \| M^{[i]}\|_{\text{op}} \le R\gamma^{i-1} \}, \label{eq:dap}
\end{align}
where $m,R$ and $\gamma$ are algorithm parameters.

\end{definition}

This class is known to be sufficiently rich to approximate many linear controllers. A policy takes in the past history and current state as input and produces a control signal as output. Let's denote $M_{1:n} := (M_1,\ldots,M_n)$ to be a sequence of DAP policies such that at time $t$, the control signal is selected using the policy parameterized by $M_t$ (see Eq.\eqref{eq:policy}). We denote $x_t^{M_{1:n}}$ to be the state reached at round $t$ by playing the sequence of policies defined by parameters $M_{1:t-1}$ in the past. Similarly $u_t^{M_{1:n}}$ is used to denote the control signal produced by the policy $M_t$. The \emph{universal dynamic regret} of the learner against the policy sequence $M_{1:n}$ is defined as:
\begin{align}
    R \left(M_{1:n} \right)
    &=\sum_{t=1}^n \ell(x_t^{\text{alg}},u_t^{\text{alg}})
    - \ell(x_t^{M_{1:n}},u_t^{M_{1:n}}), \label{eq:lqrRegret}
\end{align}
where $(x_t^{\text{alg}},u_t^{\text{alg}})$ denotes the state and control signal of the learner at round $t$. Note that the policy sequence $M_{1:n}$ can be \emph{any} valid sequence of DAP polices. The main focus of this paper is to design algorithms that can  control the dynamic regret against a sequence of reference policies as a function of the time horizon $n$ and the a path variation of the DAP parameters of the comparator $M_{1:n}$. We remark that the comparator polices $M_{1:n}$ can be chosen in hindsight and potentially unknown to the learner.

Whenever $M_{1:n} = (M,\ldots,M)$ for a fixed parameter $M$, we recover the notion of static regret. However the notion of static regret is not befitting for non-stationary environments. For example consider the scenario of controlling a drone. Suppose during the initial half of the trajectory there is heavy wind eastwards and in the second half, wind blows westwards. For best performance, a controller has to choose different policies that  can counter-act the wind and guide the motion properly in each half. Hence, we aim to control the dynamic regret which allows us to be competent against a sequence of potentially time-varying polices chosen in hindsight. We remark that our algorithm automatically adapts to the level of non-stationarity in the hindsight sequence of policies. 

Next, we take a digression and discuss a desirable property for the design of algorithms for LQR control.

\textbf{Proper learning in LQR control.} Proper learning is an online learning paradigm where the decisions of the learner are required to obey some user specified physical limits. On the other hand, improper learning framework allows the learner to disregard such constraints. The paradigm of improper learning may not be attractive in certain applications where safety is a paramount concern. Improper algorithms can possibly take the system through trajectories that are deemed to be risky. It is desirable to avoid such behaviours in physical systems such as self driving cars, control of medical ventilators, robotic control \citep{levine2016robot} and cooling data centers \citep{Cohen2018OnlineLQ}. A policy selects a control signal $u_t$ depending on the current state $x_t$. Given the value of current state, there can be physical constraints on the allowable control actions. For example, imagine the situation where we want to maintain the velocity of a drone. Depending on the current position and other system and environmental factors (the state), one can only apply a range of allowable torque (the control action) to the blades. Not respecting this torque range can drain the battery quickly or can lead to catastrophic damages such as burnt rotors. In our framework, we model this set of allowable control actions at a state as $\cF_t := \{u_t | u_t = \pi_t^M(x_t) \text{ for some } M \in \cM(m,R,\gamma) \}$ (see Definition \ref{def:dap}). So to ensure safety, at each round the learner plays a control signal from the feasible set $\cF_t$ thus necessitating the need for proper learning.

Below are our contributions:

\begin{itemize}
    \item We develop an optimal universal dynamic regret minimization algorithm for the general mini-batch linear regression problem (see Theorem \ref{thm:main}).
    
    \item Applying the reduction of \citet{Foster2020LogarithmicLQR} from LQR problem to online linear regression, the above result lends itself to an algorithm for controlling the dynamic regret of the LQR problem (Eq.\eqref{eq:lqrRegret}) to be $\tilde O^*(n^{1/3}[\mathcal{TV}(M_{1:n})]^{2/3})$, where $\mathcal{TV}$ denotes the total variation incurred by the sequence of DAP policy parameters in hindsight (see Corollary \ref{cor:lqr}). $O^*$ hides the dependencies in dimensions and system parameters.
    
    \item We show that the aforementioned dynamic regret guarantee is minimax optimal modulo dimensions and factors of $\log n$ (see Theorem \ref{thm:lb}).
    
    \item The resulting algorithm is also strongly adaptive, in the sense that the static regret against a DAP policy in any local time window is $O^*(\log n)$. 
\end{itemize}

\textbf{Notes on novelty and impact.} As discussed before, the reduction of \citet{Foster2020LogarithmicLQR} casts LQR problem to an instance of proper online linear regression.  In the context of regression, proper learning means that the decisions of the learner belongs to a user specified convex domain. The main challenge in developing aforementioned contributions rests on the design of an optimal universal dynamic regret minimization algorithm for online linear regression under the setting of \emph{proper learning}. We are not aware of any such algorithms in the literature to-date and the problem remains open. However, there exists an improper algorithm from \citet{Baby2021OptimalDR} for controlling the desired dynamic regret. Given this fact, the design of our algorithm is facilitated by coming up with \emph{new} black-box reductions (see Section \ref{sec:linr}) that can convert an improper algorithm for non-stationary online linear regression to a proper one. There are improper to proper black-box reduction schemes given in the influential work of \citet{Cutkosky2018BlackBoxRF}. However, they are developed to support general convex or strongly convex (see Definition \ref{def:sc}) losses. The linear regression losses arising in our setting are exp-concave (see Definition \ref{def:exp}) which enjoy strong curvature only in the direction of the gradients as opposed to uniformly curved strongly convex losses. Hence the reduction scheme of \citet{Cutkosky2018BlackBoxRF} is inadequate to provide fast regret rates in our setting. In contrast, we develop novel reduction schemes that carefully take the non-uniform curvature of the linear regression losses into account so as to facilitate fast dynamic regret rates (see Section \ref{sec:algo}). We remark that the algorithm ProDR.control developed in Section \ref{sec:linr} can be impactful in general online learning literature. That the non-stationary LQR problem can be \emph{optimally} solved using ProDR.control is a testament to this fact. Further our algorithm is out-of-the-box applicable to more general settings such as non-stationary multi-task linear regression, which is beyond the current scope.

The lower bound we provide in Theorem \ref{thm:lb} is also applicable to the more general problem of online non-parametric regression against a Besov space / class of Total Variation bounded functions \citep{rakhlin2014online} (see Section \ref{sec:lqr} for more details). The main contribution here is that we provide a new lower bounding strategy that characterizes the correct rate wrt both $n$ and the radius (or path-variation) of the non-parametric function class. This is in contrast with \citet{rakhlin2014online} who establish the correct dependency only wrt $n$. Deriving dependencies wrt both $n$ and the radius / path-variation is imperative in implying a dynamic regret lower bound for the LQR problem.

The rest of the paper is organized as follows. In Section \ref{sec:prelims}, we cover the necessary preliminaries on LQR control. Section \ref{sec:lit} discusses relevant literature. In Section \ref{sec:linr}, we develop a proper algorithm for non-stationary online linear regression. In Section \ref{sec:lqr}, we apply the results of Section \ref{sec:linr} to provide an algorithm for non-sationary LQR control and prove its minimax optimality. This is followed by conclusion and open problems in Section \ref{sec:concl}. We provide a concise overview of the results from \citet{baby2022optimal} in Appendix \ref{app:box} which we build upon. All proofs are given in Appendix \ref{app:proof}.

\section{Preliminaries} \label{sec:prelims}
We start with a brief overview of the LQR problem for the sake of completeness. The material of this section closely follows \citet{Foster2020LogarithmicLQR}. The definitions and notations introduced in this section will be used throughout the paper.

A linear control law is given by $u_t = -Kx_t$ for a controller $K \in \mathbb{R}^{d_u \times d_x}$. A linear controller $K$ is said to be stabilizing if $\rho(A-BK) < 1$ where $\rho(A-BK)$ is the maximum of the absolute values of the eigenvalues of $A-BK$. We assume that there exists a stabilizing controller for the system $(A,B)$. For such systems, there exists a unique matrix $P_\infty$ which is the solution to the equation:
\begin{align}
    P = A^TPA + R_x - A^TPB(R_u+B^TPB)^{-1} B^TPA.
\end{align}

The solution $P_\infty$ is called the infinite horizon Lyapunov matrix. It is an intrinsic property of the system $(A,B)$ and characterizes the optimal infinite horizon cost for control in the absence of noise \citep{bertsekas2005}. We also define the optimal state feedback controller
\begin{align}
    K_\infty := (R_u+B^TP_\infty B)^{-1} B^T P_\infty A, \label{eq:kinf}
\end{align}
the steady state covariance matrix:
\begin{align}
    \Sigma_\infty := R_u + B^TP_\infty B,
\end{align}
and the closed loop dynamics matrix: $A_{\text{cl},\infty} := A - BK_\infty$.

\citet{Foster2020LogarithmicLQR} shows that the problem of controlling the regret in the LQR problem can be reduced to online linear regression problem with delays. Specifically we have the following fundamental result due to \citet{Foster2020LogarithmicLQR}.

\begin{proposition} \label{prop:lqr}
Suppose the learner plays policy of the form $\pi_t^{\text{alg}}(x) = -K_\infty x +   q^{M_t^{\text{alg}}}(w_{1:t-1})$. Let the comparator policies take the form $\pi_t(x) = -K_\infty x +   q^{M_t}(w_{1:t-1})$ for a sequence of matrices $M_{1:n}$ chosen in hindsight. Then the dynamic regret against the policies $\pi := (\pi_1,\ldots,\pi_n)$ satisfies:
\begin{align}
    R_n(\pi)
    &\le O(1) + \sum_{t=1}^n \hat A_t(M_t^{\text{alg}},w_{t:t+h}) - \hat A_t(M_t,w_{t:t+h}),
\end{align}
where the parameters involved in the inequality are defined as below: $\hat A_t(M,w_{t:t+h}) := \|q^M(w_{1:t-1}) - q_{\infty;h}(w_{t:t+h}) \|_{\Sigma_\infty}^2$. $q_{\infty;h}(w_{t:h+t}) := \sum_{i=t+1}^{t+h} \Sigma_\infty^{-1}B^T(A_{\text{cl},\infty})^{i-1-t} P_\infty w_i$. $h := 2(1-\gamma_\infty)^{-1} \log(\kappa_\infty^2 \beta_*^2 \Psi_* \Gamma_*^2 n^2)$.  $\gamma_\infty := \|I - P+\infty^{-1/2} R_x P_\infty^{1/2}\|_{\text{op}}^{1/2}$. $\kappa_\infty := \|P_\infty^{1/2} \|_{\text{op}} \|P_\infty^{-1/2} \|_{\text{op}}$. $\beta_* := \max \{1, \lamda_{\text{min}}^{-1}(R_u), \lamda_{\text{min}}^{-1}(r_x) \}$.\\
$\Psi_* = \max\{1,\|A \|_{\text{op}},\| B\|_{\text{op}},\| R_x\|_{\text{op}},\|R_u \|_{\text{op}} \}$. $\Gamma_* := \max\{1,\| P_\infty\|_{\text{op}} \}$
\end{proposition}

Observe that the losses $\hat A_t(M,w_{t:t+h}) := \|q^M(w_{1:t-1}) - q_{\infty;h}(w_{t:t+h}) \|_{\Sigma_\infty}^2 = \hat A_t(M,w_{t:t+h}) := \|\Sigma_\infty ^{1/2} q^M(w_{1:t-1}) - \Sigma_\infty^{1/2} q_{\infty;h}(w_{t:t+h}) \|_2^2$ are essentially linear regression losses. The quantity $\Sigma_\infty ^{1/2} q^M(w_{1:t-1})$ is a linear map from the matrix sequence $M$ to $\mathbb{R}^{d_u}$. However, there is one caveat in that the bias vector at round $t$ given by $\Sigma_\infty^{1/2} q_{\infty;h}(w_{t:t+h})$ is only available at round $t+h = t + O(\log n)$. This issue of delayed feedback can be directly handled using the delayed to non-delayed online learning reduction from \citet{Joulani2013OnlineDelay}.

\section{Related work} \label{sec:lit}
In this section, we review recent progress at the intersection of control and online convex optimization (OCO) that are most relevant to our work. 

\textbf{Online control}. The idea of using tools from OCO for general control problem was proposed in \citet{agarwal2019control}. They place the notion of regret against the class of DAP policies as the central performance measure. The DAP class is also shown to be sufficiently rich to approximate a wide class of linear state-feedback controllers. Under general convex losses, they propose a reduction to OCO with memory \citep{merhav2000memory,anava2015memory} and derives $O(\sqrt{n})$ regret when the system matrices $(A,B)$ are known. For the case of unknown system, \citet{hazan2020control} provides $O(n^{2/3})$ regret via system identification techniques. When the losses are strongly convex and sub-quadratic, \citet{simchowitz2020control} strengthens these results to attain $\tilde O(n)$ regret for known systems and $\tilde O(\sqrt{n})$ when the system is unknown. For partially observable systems strong regret guarantees are provided in \citet{Simchowitz2020ImproperControl}. \citet{luo2022control} provides an $O(n^{3/5})$ dynamic regret bound for the case when the system matrices $(A_t,B_t)$ can change over time. Their results are incompatible to ours in that they consider unknown dynamics, stochastic disturbances and the dynamic regret compete with controllers that are pointwise optimal (restricted dynamic regret), while we assume known dynamics, adversarial disturbances and compete with an arbitrary sequence of controllers (i.e., universal dynamic regret).

Next, we clarify the differences from other existing work on the nonstochastic control problems \citep{gradu2020controlBandit,Gradu2020AdaptiveRF,asaf2020control,Zhang2021StronglyAO,shi2020control,Goel2020RegretoptimalCI,Zhao2021Memory}.

\citet{gradu2020controlBandit,asaf2020control} studied online control in the partially observed cases with bandit feedback, but did not consider the problem of non-stationarity with dynamic regret. \citet{Gradu2020AdaptiveRF,Zhang2021StronglyAO} studied the adaptive regret in nonstochastic control problems, which is an alternative metric to capture the performance of the learning controller in non-stationary environments. Our algorithm uses a reduction to adaptive regret too, but it is highly nontrivial to show that one can tweak adaptive regret minimizing algorithms into ones that achieve optimal dynamic regret. Moreover, our algorithm is the first that achieves logarithmic adaptive regret for nonstochastic LQR control problems too.  In contrast, \citet{Gradu2020AdaptiveRF,Zhang2021StronglyAO} focused on the slower adaptive regret in the general convex loss cases. 

To the best of our knowledge, \citet{Goel2020RegretoptimalCI} and \citet{Zhao2021Memory} are the only existing work that considered dynamic regret in non-stochastic control.

\citet{Goel2020RegretoptimalCI} used tools from $H_\infty$ control and derived a controller with exact minimax optimal dynamic regret against an oracle controller that sees the whole sequence of disturbances and chooses an optimal sequence of control actions. But the optimal dynamic regret against the sequence of control actions given by the unrealizable oracle controller is linear in $n$ in general (see an explicit lower bound from \citet{goel2020power}). It is unclear whether this oracle controller can be realized by a sequence of time-varying DAP controllers. If so, then our results would imply regret bound against the optimal sequence of control actions too. Comparing to the exact minimax regret of the $H_\infty$ style controller, our regret bound would \emph{adapt to each problem instance}, and is sublinear whenever the approximating sequence of DAP controllers has sublinear total variation.

\citet{Zhao2021Memory} studied the universal dynamic (policy) regret problem similar to ours, but works for a broader family convex loss functions. Their regret bound  $O(\sqrt{n (1 + C_n)})$ is optimal for the convex loss family. Our results show that the optimal regret improves to $\tilde{O}(n^{1/3}C_n^{2/3}\vee 1)$ when specializing to the LQR problem where the losses are quadratic. On the technical level, \citet{Zhao2021Memory} used a reduction to the dynamic regret of OCO with memory, while we reduced to the dynamic regret of OCO with delayed feedback. 

\textbf{Dynamic regret minimization in online learning}. There is a rich body of literature on dynamic regret (Eq.\eqref{eq:regOCO}) minimization. As discussed in Section \ref{sec:prelims}, the non-staionary LQR problem can be reduced to an instance of linear regression losses which are exp-concave on compact domains. There is a recent line of research \citep{Baby2021OptimalDR,baby2022optimal} that provides optimal universal dynamic regret rates under exp-concave losses. However, the algorithm of \citet{Baby2021OptimalDR} is improper, in the sense that the iterates of the learner can lie outside the feasibility set. The work of \citet{baby2022optimal} ameliorates this issue to some extend by providing proper algorithms for the particular case of $L_\infty$ constrained (box) decisions sets. The DAP policy space in Definition \ref{def:dap} is indeed not an $L_\infty$ ball. We note that if improper learning is allowed in the LQR problem, one can run the algorithms of \citet{Baby2021OptimalDR,baby2022optimal} to attain optimal dynamic regret rates. The proper learning algorithms such as \citet{zinkevich2003online,zhang2018adaptive,Cutkosky2020ParameterfreeDA,jacobsen2022free} control dynamic regret for general convex losses. However, they are not adequate to optimally minimize dynamic regret under curved losses that are strongly convex or exp-concave. The notion of restrictive dynamic regret introduced in \citet{besbes2015non} competes with a sequence of minimizers of the losses. This notion of regret can sometimes be overly pessimistic as noted in \citet{zhang2018adaptive}. There is a series of work in the direction of dynamic regret minimization in OCO such as  \citet{jadbabaie2015online,yang2016tracking,Mokhtari2016OnlineOI,chen2018non,zhang2018dynamic,goel2019OBD,arrows2019,Zhao2020DynamicRO,Zhao2021StronglyConvex,Zhao2021Memory,Chang_Shahrampour_2021}. However, to the best of our knowledge none of these works are known to attain the optimal universal dynamic regret rate for the setting of online linear regression.

\textbf{Dynamic regret for OCO vs Dynamic (Policy) regret for Control.} We emphasize that the regret in Eq.\eqref{eq:lqrRegret} is dynamic \emph{policy} regret \citep{anava2015memory,Zhao2021Memory}. The states visited by the reference policy is counterfactual and is different from that of the learner's trajectory which we observe. This is very different from the standard OCO framework where the state of both the learner and adversary are same. So bounding the policy regret seems qualitatively harder than bounding the regret in an OCO setting. Nevertheless, for the LQR problem, the fact that there exists a reduction \citep{Foster2020LogarithmicLQR} from the problem of controlling policy regret to the problem of controlling the standard OCO regret is remarkable.

\textbf{Strongly adaptive regret minimization.} There is also a complementary body of literature on strongly adaptive algorithms that focus on controlling the static regret in any local time window. For example, the algorithm of \citet{daniely2015strongly,jun2017coin} can lead to $\tilde O(\sqrt{|I|})$ static regret in any interval of $I \subseteq [n]$ under convex losses. When the losses are exp-concave the algorithm of \citet{hazan2007adaptive,koolen2016specialist,Zhang2021DualAA} can lead to $O(\log n)$ static regret in any interval.


\section{Non-stationary ``mini-batch'' linear regression}\label{sec:linr}

In view of Proposition \ref{prop:lqr}, the losses of interest are linear regression type losses. So we take a digression in this section and study the problem of controlling dynamic regret in a general linear regression setting.

\subsection{Linear regression framework}

Consider the following linear regression protocol.
\begin{itemize}
    \item At round $t$, nature reveals a co-variate matrix $A_t \in \mathbb{R}^{p \times d}$.
    
    \item Learner plays $z_t \in \cD \subset \mathbb{R}^{d}$.
    
    \item Nature reveals the loss $f_t(z) = \|A_t z - b_t\|_2^2$.
    
\end{itemize}
Under the above regression framework, we are interested in controlling the universal dynamic regret against an arbitrary sequence of predictors $u_1, \ldots, u_n \in \cD$ (abbreviated as $u_{1:n}$) :
\begin{align}
    R_n(u_{1:n}) = \sum_{t=1}^n f_t(z_t) - f_t(u_t). \label{eq:regOCO}
\end{align}
Dynamic regret is usually expressed as a function of $n$ and a path variational that captures the smoothness of the comparator sequence. We will focus on the path variational defined by: 
\begin{align}
    \mathcal{TV}(u_{1:n}) = \sum_{t=2}^n \|u_t - u_{t-1} \|_1.
\end{align}
Below are the list of assumptions made:

\textbf{Assumption 1}. Let $a_{t,i} \in \mathbb{R}^d$ be the $i^{th}$ row vector of $A_t$. We assume that $\| a_{t,i}\|_1 \le \alpha$ for all $t \in [n]$ and $i \in [p]$. Further $\| b_t\|_1 \le \sigma$ for all $t$.

\textbf{Assumption 2}. For any $x \in \cD$, $\| x\|_1 \le \chi$ and $\| x\|_\infty \le \tilde R$.

We refer this setting as mini-batch linear regression since the loss at round $t$ can be written as a sum of a batch of quadratic losses: $f_t(z) = \sum_{i=1}^p \left(z^T a_{t,i} - b_t[i] \right)^2$.

\textbf{Terminology}. For a convex loss function $f$, we abuse the notation and take $\grad f(x)$ to be a sub-gradient of $f$ at $x$. We denote $\cD_\infty(\tilde R) := \{x \in \mathbb{R}^d : \| x\|_\infty \le \tilde R \}$. 

Linear regression losses belong to a broad family of convex loss functions called exp-concave losses:

\begin{definition} \label{def:exp}
A convex function $f$ is $\alpha$ exp-concave in a domain $\cD$ if for all $x,y \in \cD$ we have $f(y) \ge f(x) + \grad f(x)^T (x-y) + \frac{\alpha}{2} (\grad f(x)^T (x-y))^2 $.
\end{definition}

The losses $f_t(z) = \|A_t z - b_t\|_2^2$ are $(2R)^{-1}$ exp-concave if $f(z) \le R$ for all $z \in \cD$ (see Lemma 2.3 in \citet{Foster2020LogarithmicLQR}).

\begin{definition} \label{def:sc}
A convex function $f$ is $\sigma$ strongly convex wrt $\| \cdot \|_2$ norm in a domain $\cD$ if for all $x,y \in \cD$ we have $f(y) \ge f(x) + \grad f(x)^T (x-y) + \frac{\sigma}{2} \|x-y\|_2^2 $.
\end{definition}

We note that if the matrix $A_t$ is rank deficient, then the losses $f_t(z)$ cannot be strongly convex. Moving forward we do not impose any restrictive assumptions on the rank of $A_t$. As mentioned in Remark \ref{rmk:rank}, the covariate matrix that arise in the reduction of the LQR problem to linear regression is not in general full rank. So we target a solution that can handle general covariate matrices irrespective of their rank.






\subsection{The Algorithm} \label{sec:algo}

\begin{figure}[h!]
	\centering
	\fbox{
		\begin{minipage}{12 cm}
		ProDR.control: Inputs - Decision set $\cD$, $G > 0$
    \begin{enumerate}
        \item At round $t$, receive $w_t$ from $\cA$.

        \item Receive co-variate matrix $A_t := [a_{t,1},\ldots,a_{t,p}]^T$.
        
        \item Play $\hat w_t \in \argmin_{x \in \cD} \max_{i=1,\ldots,p} |a_{t,i}^T ( x -  w_t)|$.
        
        \item Let $\ell_t(w) = f_t(w) + G \cdot S_t(w)$, where $f_t(w) = \|A_t w - b_t \|_2^2$ and $S_t(w) = \min_{x \in \cD} \max_{i=1,\ldots,p} |a_{t,i}^T ( x -  w)|$.
        
        \item Send $\ell_t(w)$ to $\cA$.

    \end{enumerate}
		\end{minipage}
	}
	\caption{ProDR.control: An algorithm for non-stationary and proper linear regression.}
	\label{fig:algo-multi}
\end{figure}

Starting point of our algorithm design is the work of \citet{baby2022optimal}. They provide an algorithm that attains optimal dynamic regret when the losses are exp-concave. However, their setting works only in a very restrictive setup where the decision set is an $L_\infty$ constrained box. Consequently, we cannot directly apply their results to the linear regression problem of Section \ref{sec:linr} whenever the decision set $\cD$ is a general convex set. 

An online learner is termed proper if the decisions of the learner are guaranteed to lie within the feasibility set $\cD$. Otherwise it is called improper. A recent seminal work of \citet{Cutkosky2018BlackBoxRF} proposes neat reductions that can convert an improper online learner to a proper one, whenever the losses are convex.  Following this line of research, we can aim to convert the algorithm of \citet{baby2022optimal} that works exclusively on box decision set to one that can support arbitrary convex decision sets by coming up with suitable reduction schemes. However, the specific reduction scheme proposed in \citet{Cutkosky2018BlackBoxRF} is inadequate to yield fast dynamic rates for exp-concave losses. Our algorithm ProDR.control (Fig.\ref{fig:algo-multi}, \textbf{Pro}per \textbf{D}ynamic \textbf{R}egret.control) is a by-product of constructing new reduction schemes to circumvent the aforementioned problem for the case of linear regression losses. We expand upon these details below.

In ProDR.control, we maintain a surrogate algorithm $\cA$, which is chosen to be the algorithm of \citet{baby2022optimal} that produces iterates $w_t$ in an $L_\infty$ norm ball (box), $\cD_\infty$, that encloses the actual decision set $\cD$. Since $w_t$ can be infeasible, we play $\hat w_t$ obtained via a special type of projection of $w_t$ onto $\cD$ which is formulated as a min-max problem in Line 3 of Fig.\ref{fig:algo-multi}. In Line 4, we construct surrogate losses $\ell_t$ to be passed to the algorithm $\cA$. The surrogate loss penalises $\cA$ for making predictions outside $\cD$. We will show (see Lemma \ref{lem:sur} in Appendix) that the instantaneous regret satisfies $f_t(\hat w_t) - f_t(u_t) \le \ell_t(w_t) - \ell_t(u_t)$, where $u_t \in \cD$ is the comparator at round $t$. Thus the dynamic regret of the proper iterates $\hat w_t$ wrt linear regression losses is upper bounded by the dynamic regret of the surrogate algorithm $\cA$ on the losses $\ell_t$ and box decision set.

The design of the min-max barrier $S_t(w)$ is driven to ensure exp-concavity of the surrogate losses $\ell_t(w) = f_t(w) + G\cdot S_t(w)$. We capture its intuition as follows. We start by observing that since $\grad^2 f_t(w) = 2A_t^TA_t$, the linear regression losses $f_t$ exhibits strong curvature along the row-space of $A_t$, denoted by $\row(A_t)$. Further we have $\grad f_t(w) = 2 A_t^T(A_t w - b_t) \in \row(A_t)$. So the loss $f_t$ exhibits strong curvature along the direction of its gradient too. This is the fundamental reason behind the exp-concavity of $f_t$. The min-max barrier $S_t(w)$ is designed such that its gradient is guaranteed to lie in the $\row(A_t)$ (see Lemma \ref{lem:subg} in Appendix for a formal statement). So the overall gradient $\grad \ell_t(w)$ also lies in the $\row(A_t)$. Since the function $f_t$ already exhibits strong curvature along $\row(A_t)$, we conclude that the sum $\ell_t(w) = f_t(w) + G\cdot S_t(w)$ exhibits strong curvature along its gradient $\grad \ell_t(w)$. This maintains the exp-concavity of the losses $\ell_t$ over $\cD_\infty$ (see Lemma \ref{lem:exp} in Appendix). Such curvature considerations along with the fact that $S_t(w)$ has to be sufficiently large to facilitate the instantaneous regret bound $f_t(\hat w_t) - f_t(u_t) \le \ell_t(w_t) - \ell_t(u_t)$ results in functional form for $S_t(w)$ displayed in Fig.\ref{fig:algo-multi}.

Consequently the fast dynamic regret rates derived in \citet{baby2022optimal} becomes directly applicable. The reduction scheme used by \citet{Cutkosky2018BlackBoxRF} for producing proper iterates $\hat w_t$ and their accompanying surrogate loss design $\ell_t$ also allows one to upper bound the regret wrt linear regression losses $f_t$ by the regret of the algorithm $\cA$ wrt surrogate losses $\ell_t$. However, the surrogate loss $\ell_t$ they construct is not guaranteed to be exp-concave and consequently not amenable to fast dynamic regret rates. 

\subsection{Main Results}
We have the following guarantee for ProDR.control:
\begin{restatable}{theorem}{thmmain} \label{thm:main}
Let $u_{1:n} \in \cD$ be any comparator sequence. In Fig.\ref{fig:algo-multi}, choose $G$ such that $\sup_{w_1,w_2 \in \cD_\infty(\tilde R), t\in [n]}  \|A_t (w_1+ w_2) - 2b_t \|_1 \le G$. Let $\alpha$ be as in Assumption 2. Let $L$ be such that $\sup_{w \in \cD_\infty(\tilde R), j \in [p]} 2\|A_tw -b_t\|_2^2+2G^2 \le L$ for all $t \in [n]$. Choose $\cA$ as the algorithm from \citet{baby2022optimal} (see Appendix \ref{app:box}) with parameters $\gamma = 2G\alpha \tilde R\sqrt{d/8L} + \sqrt{2L}$ and $\zeta = \min\{ \frac{1}{16G\alpha \tilde R \sqrt{d}}, 1/(4\gamma^2)\}$ and decision set $\cD_\infty(\tilde R)$. Under Assumptions 1 and 2, a valid of assignment of $G$ and $L$ are $2p\chi + 2\sigma$ and $6(p \chi + \sigma)^2$ respectively.

Then the algorithm ProDR.control yields a dynamic regret rate of 
\begin{align}
    \sum_{t=1}^n f_t(\hat w_t) - f_t(u_t)
    &= \tilde O(d^3n^{1/3}[\mathcal{TV}(u_{1:n})]^{2/3} \vee 1),
\end{align}
where $(a \vee b) := \max \{a , b \}$. 
\end{restatable}

\begin{remark}
In view of Proposition 10 in \citet{Baby2021OptimalDR}, the dynamic regret guarantee in Theorem \ref{thm:main} is optimal modulo dependencies in $d$ and $\log n$. Further the algorithm does not require apriori knowledge of the path length $\mathcal{TV}(u_{1:n})$.
\end{remark}
\begin{proof}[\textbf{Proof sketch for Theorem \ref{thm:main}}.] First step is to show that $f_t(\hat w_t) \le \ell_t(w_t)$. This is accomplished by Lipschitzness type arguments. For any $u \in \cD$, one observes that $\ell_t(u) = f_t(u)$. So the instantaneous regret of ProDR.control, $f_t(\hat w_t) - f_t(u_t)$, is upper bounded by the instantaneous regret, $\ell_t(w_t) - \ell_t(u_t)$ of the surrogate algorithm $\cA$. The crucial step is to show the exp-concavity of the losses $\ell_t$ across $\cD_\infty(\tilde R)$. For this, we prove that there is a sub-gradient $\grad S_t(w)$ that is aligned with $a_{t,j}$ for some $j \in [p]$. This observation followed by few algebraic manipulations (see proof of Lemma \ref{lem:exp} in Appendix) allows us to show the exp-concavity of $\ell_t$ over $\cD_\infty(\tilde R)$. Now the overall regret can be controlled if the surrogate algorithm $\cA$ provides optimal dynamic regret under exp-concave losses and box decision sets, $\cD_\infty(\tilde R)$. This is accomplished by choosing $\cA$ as the algorithm in \citet{baby2022optimal} which is also strongly adaptive.
\end{proof}

Since the surrogate algorithm $\cA$ we used in Theorem \ref{thm:main} is strongly adaptive (see for eg. Appendix \ref{app:box}), we also have the following performance guarantee in terms of static regret:

\begin{proposition} \label{prop:sa}
Consider the instantiation of ProDR.control in Theorem \ref{thm:main}. Then for any time window $[a,b] \subseteq [n]$ we have that: $\sum_{t=a}^b f_t(\hat w_t) - \inf_{u \in \cD} \sum_{t=a}^bf_t(u) = \tilde O(d^{1.5} \log n)$.
\end{proposition}

\begin{remark}
Theorem \ref{thm:main} and Proposition \ref{prop:sa} together makes the algorithm ProDR.control a good candidate for performing proper online linear regression in non-stationary environments.
\end{remark}

\subsection{Linear regression with delayed feedback} \label{sec:delay}
In this section, we consider a linear regression protocol with feedback delayed by $\tau$ time steps.

\begin{itemize}
    \item At round $t$, nature reveals a co-variate matrix $A_t \in \mathbb{R}^{p \times d}$.
    
    \item Learner plays $z_t \in \cD \subset \mathbb{R}^{d}$.
    
    \item Nature reveals the loss $f_{t-\tau+1}(z) = \|A_{t-\tau+1} z - b_{t-\tau+1}\|_2^2$.
    
\end{itemize}

This delayed setting can be handled by the framework developed in \citet{Joulani2013OnlineDelay}. Although these authors focus on bounding the regret as a function of time horizon $n$, the extension to dynamic regret bounds expressed in terms of both $n$ and $\mathcal{TV}(u_{1:n})$ can be handled straight-forwardly in the analysis. We include the analysis in Appendix \ref{app:proof} for the sake of completeness. The entire algorithm is as shown in Fig.\ref{fig:algo-delay}.

\begin{figure}[h!]
	\centering
	\fbox{
		\begin{minipage}{12 cm}
		ProDR.control.delayed: Inputs- delay $\tau > 0$

\begin{itemize}
    \item Maintain $\tau$ separate instances of ProDR.control (Fig.\ref{fig:algo-multi}). Enumerate them by $0,1,\ldots,\tau-1$.
    \item At time $t$:
    \begin{enumerate}
        \item Update instance $(t-1) \mod \tau$ with loss $f_{t-\tau}$.
        \item Predict using instance $(t-1) \mod \tau$.
    \end{enumerate}
\end{itemize}

		\end{minipage}
	}
	\caption{ProDR.control.delayed: An instance of delayed to non-delayed reduction from \citet{Joulani2013OnlineDelay}}
	\label{fig:algo-delay}
\end{figure}

We have the following regret guarantee for Algorithm ProDR.control.delayed.

\begin{restatable} {theorem} {thmdelay} \label{thm:delay}
Let $x_t$ be the prediction of the algorithm in Fig. \ref{fig:algo-delay} at time $t$. Instantiating each ProDR.control instance by the parameter setting described in Theorem \ref{thm:main}. Let $\tau$ be the feedback delay. We have that
\begin{align}
    \sum_{t=1}^n f_t(x_t) - f_t(u_t)
    &= \tilde O(d^3\tau^{2/3}n^{1/3}[\mathcal{TV}(u_{1:n})]^{2/3} \vee \tau).
\end{align}
Further for any interval $[a,b] \subseteq [n]$:
\begin{align}
    \sum_{t=a}^b f_t(x_t) - f_t(u) = O(d^{1.5} \tau \log n).
\end{align}

\end{restatable}

\section{Instantiation for the LQR problem} \label{sec:lqr}
In view of Proposition \ref{prop:lqr}, the LQR problem is reduced to a mini-batch linear regression problem with delayed feedback, where the delay is given by $h = O(\log n)$ in Proposition \ref{prop:lqr}. In this section, we provide explicit form of the linear regression losses arising in the LQR problem and instantiate Algorithm ProDR.control.delayed (Fig.\ref{fig:algo-delay}). First we need to define certain quantities:

For a sequence of matrices $(M^{[i]})_{i=1}^m$ define $\texttt{flatten}((M^{[i]})_{i=1}^m)$ as follows:
Let $M_k^{[i]}$ be the $k^{th}$ column of $M^{[i]}$.

Let's define 
\begin{gather}
    z^k
    =
    \begin{bmatrix}
    M^k_1\\
    \vdots \\
    M^k_{d_x}
    \end{bmatrix} \in \mathbb{R}^{d_u d_x},
\end{gather}
and
\begin{gather}
    \texttt{flatten}((M^{[i]})_{i=1}^m)
    :=
    \begin{bmatrix}
    z^1\\
    \vdots\\
    z^m
    \end{bmatrix} \in \mathbb{R}^{md_u d_x}. \label{eq:M}
\end{gather}

For a sequence of DAP parameters $M_{1:n}$, let $\mathcal{TV}(M_{1:n}):= \sum_{t=2}^{n}\sum_{i=1}^m \| M_t^{[i]} - M_{t-1}^{[i]}\|_1$.  We define $\texttt{deflatten}$ as the natural inverse operation of $\texttt{flatten}$. We have the following Corollary of Theorem \ref{thm:delay} and Proposition \ref{prop:lqr}.

\begin{restatable} {corollary} {corlqr}\label{cor:lqr} Assume the notations in Fig.\ref{fig:algo-multi} and Section \ref{sec:prelims}. Let $\Sigma_\infty = U_\infty^T \Lambda_\infty U_\infty$ be the spectral decomposition of the positive semi definite (PSD) matrix $\Sigma_\infty \in \mathbb{R}^{d_u \times d_u}$. . Let the covariate matrix $A_t := [w_{t-1}^T \ldots w_{t-m}^T] \otimes \Lambda_\infty^{1/2} U_\infty \in \mathbb{R}^{d_u \times m d_u d_x}$, where $\otimes$ denotes the Kronecker product. Let the bias vector $b_t:=\Lambda_\infty^{1/2} U_\infty q_{\infty;h}^*(w_{t:t+h})$. Let the delay factor of ProDR.control.delayed (Fig.\ref{fig:algo-delay}) be $\tau = h$ as defined in Proposition \ref{prop:lqr} and let the decision set given to the ProDR.control instances in Fig.\ref{fig:algo-delay} be the DAP space defined in Eq.\eqref{eq:dap}. Let $z_t$ be the prediction at round $t$ made by the ProDR.control.delayed algorithm and let $M_t^{\text{alg}} := \texttt{deflatten}(z_t)$. At round $t$, we play the control signal $u_t^{\text{alg}}(x_t) = \pi_t^{M_t^{\text{alg}}}(x_t)$ according to Eq.\eqref{eq:policy}. There exists a choice of input parameters for the ProDR.control instances in Fig.\ref{fig:algo-delay} such that 
\begin{align}
    R(M_{1:n})
    &=\sum_{t=1}^n \ell(x_t^{\text{alg}},u_t^{\text{alg}})
    - \ell(x_t^{M_{1:n}},u_t^{M_{1:n}})\\
    &= \tilde O \left(m^3 d^4 d_x^5 (d_u \wedge d_x) (n^{1/3} [\mathcal{TV}(M_{1:n})]^{2/3} \vee 1)\right),
\end{align}
where $M_{1:n}$ is a sequence of DAP policies where each $M_t \in \cM$ (eq.\eqref{eq:dap}). Further the algorithm ProDR.control.delayed also enjoys a strongly adaptive regret guarantee for any interval $[a,b] \subseteq [n]$:
\begin{align}
    \sum_{t=a}^b \ell(x_t^{\text{alg}},u_t^{\text{alg}}) - \ell(x_t^{M},u_t^{M})
    &= \tilde O((md_ud_x)^{1.5} \log n),
\end{align}
for any fixed DAP policy $M \in \cM$.

\end{restatable}


The following theorem provides a nearly matching lower bound.
\begin{restatable}{theorem}{thmlb} \label{thm:lb}
There exists an LQR system, a choice of the perturbations $w_t$ and a DAP policy class such that:
\begin{align}
    \sup_{M_{1:n} \text{ with } \mathcal{TV}(M_{1:n}) \le C_n} \E[R(M_{1:n})]
    &= \Omega(n^{1/3}C_n^{2/3} \vee 1),
\end{align}
where the expectation is taken wrt randomness in the strategies of the agent and adversary.
\end{restatable}
The proof of the above lower bound given in Appendix \ref{app:proof} is interesting in its own right. The proof is also applicable to the problem of online non-parametric regression against Total Variation (TV) bounded sequences \citep{rakhlin2014online,Baby2021OptimalDR}. The lower bounding strategy in \citet{rakhlin2014online} goes through arguments based on sequential Rademacher complexity of the non-parametric class of TV bounded sequences. While they establish the rate wrt $n$ as $n^{1/3}$, the correct dependency on the TV of the sequence was not provided in \citet{rakhlin2014online}. The work of \citet{Baby2021OptimalDR} ameliorated this issue by appealing to the standard lowerbounds from \emph{offline} non-parametric regression literature. This lower bounding route uses fairly sophisticated arguments based on characterizing the Bernstein width of the set of Haar wavelet coefficients of TV bounded sequences \citep{donoho1990minimax}. In contrast, we provide a lower bound capturing the correct rate wrt both $n$ and TV of the sequence via more direct arguments based on constructing an explicit adversarial strategy. An elaborate outline of applying our lower bound to the online non-parametric regression framework is given in Appendix \ref{app:proof}.

\begin{remark} \label{rmk:rank}
The covariate matrix $A_t \in \mathbb{R}^{d_u \times m d_u d_x}$ that arise in Corollary \ref{cor:lqr} is rank deficient whenever $md_x > 1$. In such cases, the linear regression losses $f_t(w)$ as in Fig.\ref{fig:algo-multi} cannot be strongly convex. So the proper universal dynamic regret minimizing algorithm for strongly convex losses from \citet{baby2022optimal} is inapplicable in general except potentially for the particular setting of $m=d_x=1$. Moreover, in the setting of $m=d_x=1$ a non-zero strong convexity parameter can exist only if the magnitude of the perturbations $|w_t|$ are bounded away from zero which is restrictive in its scope.
\end{remark}

\section{Conclusion} \label{sec:concl}
In this paper, we designed a new algorithm for online non-stochastic LQR controller.  The controller provably minimizes the regret with any oracle non-stationary sequence of Disturbance Action controllers chosen to handle the sequence of adversarially-chosen disturbances after they realized. We also show that no other algorithm is able to have smaller max-regret by more than a logarithmic factor, i.e., our proposed algorithm is optimal. The underlying algorithm is a new development in minimizing dynamic regret in non-stationary (minibatched) linear regression problem under the \emph{proper} learning setup. The techniques developed in this work can be of independent interest in the broader literature of online learning. Future work include generalizing the family of loss functions to general strongly convex losses and exp-concave losses.

\section*{Acknowledgements}
The construction of the lower bound in Theorem \ref{thm:lb} is due to an early discussion with Daniel Lokshtanov on a related problem.


\bibliography{tf,yx}
\bibliographystyle{plainnat}

\newpage 

\appendix

\section{Brief overview of results from \citet{baby2022optimal}} \label{app:box}
For the sake of completeness, we dedicate this session for a short discussion about the results of \citet{Baby2021OptimalDR}.

First, we recall the description of Follow-the-Leading-History (FLH) algorithm from \citep{hazan2007adaptive}.

\begin{figure}[h!]
	\centering
	\fbox{
		\begin{minipage}{12 cm}
		FLH: inputs - Learning rate $\zeta$ and $n$ base learners $E^1,\ldots,E^n$
            \begin{enumerate}
                \item For each $t$, $v_t = (v_t^{(1)},\ldots,v_t^{(t)})$ is a probability vector in $\mathbb{R}^t$. Initialize $v_1^{(1)} = 1$.
                \item In round $t$, set $\forall j \le t$, $x_t^j \leftarrow E^j(t)$ (the prediction of the $j^{th}$ bas learner at time $t$). Play $x_t =  \sum_{j=1}^t v_t^{(j)}x_t^{(j)}$.
                \item After receiving $f_t$, set $\hat v_{t+1}^{(t+1)} = 0$ and perform update for $1 \le i \le t$:
                \begin{align}
                    \hat v_{t+1}^{(i)}
                    &= \frac{v_t^{(i)}e^{-\zeta f_t(x_t^{(i)})}}{\sum_{j=1}^t v_t^{(j)}e^{-\zeta f_t(x_t^{(j)})}}
                \end{align}
                \item Addition step - Set $v_{t+1}^{(t+1)}$ to $1/(t+1)$ and for $i \neq t+1$:
                \begin{align}
                    v_{t+1}^{(i)}
                    &= (1-(t+1)^{-1}) \hat v_{t+1}^{(i)}
                \end{align}
            \end{enumerate}
		\end{minipage}
	}
	\caption{FLH algorithm}
	\label{fig:flh}
\end{figure}

Next, we describe Online Newton Step (ONS) algorithm from \citet{hazan2007logregret}.

\begin{figure}[h!]
	\centering
	\fbox{
		\begin{minipage}{12 cm}
		ONS: inputs - $\zeta$. Decision set $\cD$.
            \begin{enumerate}
            \item At round 1, predict $0$.
            \item At iteration $t >1$ predict:
            \begin{align}
            w_t \in \argmin_{x \in \cD} \|w_{t-1} - \frac{1}{\beta} A_{t-1}^{-1} \grad_{t-1} - x \|_{A_{t-1}},
            \end{align}
            where $\grad_\tau = \grad f_\tau(x_\tau),\: A_t = \zeta I_d + \sum_{i=1}^t \grad_i \grad_i^{\top}$.
            \end{enumerate}
		\end{minipage}
	}
	\caption{ONS algorithm}
	\label{fig:ons}
\end{figure}

\textbf{Assumption A1:} The loss functions $\ell_t$ are $\alpha$ exp-concave in the box decision set $\cD=\{x \in \mathbb{R}^d: \| x\|_\infty \le B \}$ .ie, $\ell_t(y) \ge \ell_t(x) + \grad \ell_t(x)^T (y- x)+ \frac{\alpha}{2} \left( \grad \ell_t(x)^T (y- x) \right)^2$ for all $x, y \in \cD$.

\textbf{Assumption A2:} The loss functions $\ell_t$ satisfy $\| \grad \ell_t(x)\|_2 \le G$ and $\| \grad \ell_t(x)\|_\infty \le G_\infty$ for all $x \in \cD$. Without loss of generality, we let $G \wedge G_\infty \wedge B \ge 1$, where $a \wedge b := \min \{a , b \}$.

We consider the following protocol:
\begin{itemize}
    \item At time $t \in [n]$ learner predicts $x_t \in \mathbb{R}^d$ with $\| x_t\|_\infty \le B$.
    \item Adversary reveals the loss function $\ell_t$.
\end{itemize}

In view of Assumption A1, following \citep{hazan2007logregret}, one can define the surrogate losses:
\begin{align}
    f_t(x) = \left(\sqrt{\alpha/2} \grad \ell_t(x_t)^T (x- x_t) + 1/\sqrt{2\alpha} \right)^2. \label{eq:exp-surrogate}
\end{align}

The surrogate losses satisfy the following property:
\begin{align}
    \sum_{t=1}^n \ell_t(x_t) - \ell_t(w_t)
    &\le \sum_{t=1}^n f_t(x_t) - f_t(w_t),
\end{align}
where $x_t, w_t \in \cD$.

We have the following dynamic regret guarantee from \citet{Baby2021OptimalDR}.

\begin{restatable}{theorem}{main}\label{thm:ec-d}
Suppose Assumptions A1-A2 are satisfied. Define $\gamma := 2GB\sqrt{\alpha d/2} + 1/\sqrt{2 \alpha}$. By using the base learner as ONS with parameter $\zeta = \min \left \{\frac{1}{16GB\sqrt{d}}, 1/(4 \gamma^2) \right \}$, decision set $\cD$, loss at time $t$ to be $f_t$ and  choosing learning rate of FLH as $\eta = 1/(2 \gamma^2)$, FLH-ONS obeys
\begin{align}
    \sum_{t=1}^n \ell_t(x_t) - \ell_t(w_t)
    &\le \tilde O \left( 140 d^2(8G^2B^2\alpha d + G^2B^2 + 1/\alpha) (n^{1/3}[\mathcal{TV}(u_{1:n})]^{2/3} \vee 1) \right)\mathbb{I}\{ \mathcal{TV}(u_{1:n}) > 1/n\}\\
    &\quad + \tilde O\left( d(8G^2B^2\alpha d + 1/\alpha \right) \mathbb{I}\{ \mathcal{TV}(u_{1:n}) \le 1/n\},
\end{align}
where $x_t$ is the decision of the algorithm at time $t$ and $\tilde O(\cdot)$ hides polynomial factors of $\log n$. $\mathbb{I}\{\cdot \}$ is the boolean indicator function assuming values in $\{0,1 \}$.
\end{restatable}

We also have the following strongly adaptive regret guarantee:
\begin{theorem}
Consider the setting of FLH-ONS in Theorem \ref{thm:ec-d}. Then for any interval $[a,b] \subseteq [n]$ we have that
\begin{align}
     \sum_{t=a}^b \ell_t(x_t) - \ell_t(w_t)
     &= O(d^{1.5} \log n).
\end{align}
\end{theorem}

\section{Omitted Proofs} \label{app:proof}

In this section we use the notations defined in Fig.\ref{fig:algo-multi}.

The lemma below shows how the surrogate losses $\ell_t$ can be used to upper bound the regression losses $f_t$.

\begin{lemma} \label{lem:sur}
Assume the notations in Fig.\ref{fig:algo-multi}. Let $G$ be such that $\sup_{w_1,w_2 \in \cD_\infty(\tilde R)}  \|A_t (w_1+ w_2) - 2b_t \|_1 \le G$ for all $t \in [n]$. We have that:
\begin{itemize}
    \item $f_t(\hat w_t) \le \ell_t(w_t)$,
    
    \item $f_t(u) = \ell_t(u)$ for all $u \in \cD$
\end{itemize}

\end{lemma}
\begin{proof}

For any $w_1, w_2 \in \cD_\infty(\tilde R)$
\begin{align}
    f_t(w_1) - f_t(w_2)
    &= \|A_t w_1 - b_t \|_2^2 - \|A_t w_2 - b_t \|_2^2\\
    &= \left (A_t (w_1+ w_2) - 2b_t \right)^T \left ( A_t (w_1 -w_2) \right)\\
    &\le \|A_t (w_1+ w_2) - 2b_t \|_1 \|A_t (w_1 -w_2) \|_\infty\\
    &\le G \max_{i=1,\ldots,p} |a_{t,i}^T (w_1 - w_2)|, \label{eq:lip-main}
\end{align}

for a $G$ such that $\sup_{w_1,w_2 \in \cD_\infty(\tilde R)}  \|A_t (w_1+ w_2) - 2b_t \|_1 \le G$ holds true.

In particular we have that:
\begin{align}
    f_t(\hat w_t)
    &\le f_t(w_t) + G \max_{i=1,\ldots,p} |a_{t,i}^T (\hat w_t - w_t)| := \ell_t(w_t)
\end{align}

For any $u \in \cD$, we have that $S_t(u) = 0$. Hence $f_t(u) = \ell_t(u)$.

\end{proof}

The lemma below establishes certain useful properties of the barrier function $S_t(w)$.

\begin{restatable}{lemma}{lemsubg} \label{lem:subg}
The function $S_t(w)$ satisfies the following properties:

\begin{enumerate}
    \item $S_t(w) = \max_{i=1,\ldots,p} \min_{x \in \cD} |a_{i,t}^T(x-w)|$.

    \item $S_t(w)$ is convex over $\mathbb{R}^d$.

    \item Let $i^*$ be such that  $S_t(w) = \min_{x \in \cD} |a_{i^*,t}^T(x-w)|$. Let $\Pi(w) \in \argmin_{x \in \cD} |a_{i^*,t}^T(x-w)|$. Let $g_t \in \partial S_t(w)$, When $a_{i^*,t}^T(\Pi(w)-w) \neq 0$ we have:

	\begin{equation*}
	g_t = \left\{
	\begin{aligned}
	a_{i^*,t},\quad&\text{ if }\quad a_{i^*,t}^{\top}(\Pi(w)-w) < 0\\
	-a_{i^*,t},\quad&\text{ if }\quad   a_{i^*,t}^{\top}(\Pi(w)-w) > 0.
	\end{aligned}
	\right.
	\end{equation*}    
	
	If $a_{i^*,t}^T(\Pi(w)-w) = 0$ then we take $g_t = 0$.
    
\end{enumerate}
\end{restatable}
\begin{proof}
We set out to prove the first statement. Let $\Delta_p$ be the $p$ dimensional simplex. We have that
\begin{align}
    S_t(w) 
    &= \min_{x \in \cD} \max_{i=1,\ldots,p} |a_{i,t}^T(x-w)|\\
    &=_{(a)} \min_{x \in \cD} \max_{v \in \Delta_p} \sum_{i=1}^p v_i |a_{i,t}^T(x-w)|\\
    &=_{(b)} \max_{v \in \Delta_p} \min_{x \in \cD} \sum_{i=1}^p v_i |a_{i,t}^T(x-w)|.
\end{align}
For line (a) we observed that for a given $x$ $\max_{v \in \Delta_p} \sum_{i=1}^p v_i |a_{i,t}^T(x-w)|$ is attained by putting all the weights of $v$ to an $i^* \in \argmax_{i=1,\ldots,p}  |a_{i,t}^T(x-w)|$.

For line (b) we observe that the function $r(x,v) = \sum_{i=1}^p v_i |a_{i,t}^T(x-w)|$ is a convex function of $x$ and concave function of $p$. So by applying Sion's minimax theorem we arrive at line (b).

Next we set out to prove that:
\begin{align}
    \max_{v \in \Delta_p} \min_{x \in \cD} r(x,v)
    &= \max_{i=1,\ldots,p} \min_{x \in \cD} |a_{i,t}^T (x-w)|\label{eq:relax}
\end{align}

Let $(x^*,v^*)$ be a solution that attains $\max_{v \in \Delta_p} \min_{x \in \cD} r(x,v)$. Further, for the sake of contradiction, let's assume that $v^* \neq e_k$ for any $k \in [p]$. ($e_k$ is the unit vector with $1$ at entry $k$). Let the index $j$ be such that $|a_{j,t}^T (x^*-w)| > |a_{i,t}^T (x^*-w)|$ for all $i \in [p] \setminus \{ j\}$. Then we can find a solution $e_{j}$  such that  $r(x^*,e_j) > r(x^*,v^*)$. This contradicts the fact that $(x^*,v^*)$ is a valid solution.

In the alternate case let $j$ be an index in $[p]$ such that $|a_{j,t}^T (x^*-w)| \ge |a_{i,t}^T (x^*-w)|$ for all $i \in [p] \setminus \{ j\}$. Suppose for all $ i \in Q \subseteq [p] \setminus \{ j\}$ we have $|a_{j,t}^T (x^*-w)| = |a_{i,t}^T (x^*-w)|$. By earlier arguments, we must have $v^*[k]$ must be equal to zero for all $k \in [p] \setminus (Q \cup \{ j\})$.  Then putting all the weight to $j$ produces an equally valid solution in the sense that $r(x^*,e_j) =r(x^*,v^*) $

Combining the above two cases, we conclude that there exists maximizers $v^*$ such that $v^* = e_k$ for some $k \in [p]$. This leads to Eq.\eqref{eq:relax}.

Next we prove statement 2. For any given $i$ we have that $|a_{i,t}^T(x-w)|$ is a convex function of both $x$ and $w$. Hence the point-wise maximum $\max_{i=1,\ldots,p}  |a_{i,t}^T(x-w)|$ is also convex in both $x$ and $w$. Since partial minimisation preserves convexity, we have that $\min_{x \in \cD} \max_{i=1,\ldots,p}  |a_{i,t}^T(x-w)|$ remains convex in $w \in \mathbb{R}^d$.

Next we prove statement 3. We know that sub-gradient set of point-wise maximum of convex functions is the convex hull of sub-gradients of the active functions. Applying this result along with the sub-gradient characterization of the function $\min_{x \in \cD} |a_{i,t}^T (x-w)|$ in Lemma \ref{lem:subg-full} leads to the third statement.

\end{proof}

The next lemma establishes the exp-concavity of the surrogate losses $\ell_t$ over the decision domain of the surrogate algorithm $\cA$.

\begin{lemma} \label{lem:exp}
Assume the notations in Fig.\ref{fig:algo-multi}. Let $L$ be such that $\sup_{w \in \cD_\infty(\tilde R), j \in [p]} 2\|A_tw -b_t\|_2^2+2G^2 \le L$ for all $t \in [n]$. Then the losses $\ell_t$ are  exp-concave over $\cD_\infty(\tilde R)$ with parameter $1/4L$.
\end{lemma}
\begin{proof}
Observe that $\grad f_t(w) = 2 A_t^T(A_t w - b_t)$ and $\grad^2 f_t(w) = 2A_t^TA_t$.

We have that for any $w_1, w_2 \in \mathbb{R}^d$
\begin{align}
    f_t(w_2) = f_t(w_1) + \langle \grad f_t(w_1), w_2 - w_1 \rangle + \frac{1}{2} \| w_2 - w_1\|^2_{2A_t^TA_t}. \label{eq:m1}
\end{align}

Due to the convexity of $S_t(w)$ over $\mathbb{R}^d$ from Lemma \ref{lem:subg}, we have that
\begin{align}
    S_t(w_2) \ge S(w_1) + \langle \grad S_t(w_1), w_2 -w_1 \rangle. \label{eq:m2}
\end{align}

Combining Eq.\eqref{eq:m1} and \eqref{eq:m2} we have that
\begin{align}
    \ell_t(w_2) \ge \ell_t(w_1) + \langle \grad \ell_t(w_1), w_2 - w_1 \rangle + \frac{1}{2} \| w_2 - w_1\|^2_{2A_t^TA_t}
\end{align}

Observe that $\grad \ell_t(w_1) = 2A_t^T(A_tw_t-b_t) + Gh A_t^T e_j$, for some $h \in \{-1,0,1\}$ and $j \in [p]$ due to Lemma \ref{lem:subg}. Now, let's focus on points $w_1,w_2 \in \cD_\infty(\tilde R)$. We have

\begin{align}
    \grad \ell_t(w_1) \grad \ell_t(w_1)^T 
    &= 4A_t^T(A_tw_1-b_t+Gh e_j)(A_tw_1-b_t+Gh e_j)^TA_t\\
    &\preccurlyeq 4LA_t^TA_t,
\end{align}
$L$ is such that:
\begin{align}
 \sup_{w \in \cD_\infty(\tilde R), j \in [p]} \|(A_tw - b_t+Gh e_j) \|_2^2 \le L   . \label{eq:L}
\end{align}

Hence for all $w_1,w_2 \in \cD_\infty(\tilde R)$, we have the relation
\begin{align}
    \ell_t(w_2) \ge \ell_t(w_1) + \langle \grad \ell_t(w_1), w_2 - w_1 \rangle + \frac{1}{4L} \| w_2 - w_1\|^2_{\grad \ell_t(w_1) \grad \ell_t(w_1)^T}.
\end{align}

Thus the losses $\ell_t$ remains exp-concave over $\cD_\infty(\tilde R)$ with parameter $1/4L$.

\end{proof}

We are now ready to prove Theorem \ref{thm:main}.
\thmmain*
\begin{proof}
From Eq.\eqref{eq:lip-main} we have that for any $w_1,w_2 \in \cD_\infty(\tilde R)$
\begin{align}
    f_t(w_1) - f_t(w_2)
    &\le G\alpha \|w_1 - w_2 \|_2, 
\end{align}

for a $G$ such that $\sup_{w_1,w_2 \in \cD_\infty(\tilde R)}  \|A_t (w_1+ w_2) - 2b_t \|_1 \le G$ holds true.

From Lemma \ref{lem:subg} we have for any subgradient $\| \grad S_t(w) \|_2 \le \alpha$ (where $\alpha$ is as in Assumption 1). Thus the losses $\ell_t$ are $2G\alpha$-Lipschitz in L2 norm over $\cD_\infty(\tilde R)$. Now combining Lemma \ref{lem:exp} and Theorem 10 in \citet{baby2022optimal} (or see Appendix \ref{app:box}) we have that
\begin{align}
    \sum_{t=1}^n \ell_t(w_t) - \ell_t(u_t)
    &= \tilde O\left((d^3G^2\alpha^2\tilde R^2/L + d^2G^2\alpha^2\tilde R^2 + d^2L) (n^{1/3}[\mathcal{TV}(u_{1:n})]^{2/3} \vee 1) \right))\\
    &= \tilde O(d^3n^{1/3}[\mathcal{TV}(u_{1:n})]^{2/3} \vee 1).
\end{align}

Applying Lemma \ref{lem:sur} now concludes the proof.
\end{proof}

\begin{lemma} \label{lem:subg-full}
Let $f(x) = min_{u \in \cD} |a^T(u-x)|$ for a compact and convex set $\cD$. Let $0 \in \cD$. f(x)  is convex. Let $s \in argmin_{u \in \cD} |a^T(u-x)|$. 
\begin{align}
    \grad f(x)
    &= \begin{cases}
    -a & a^T(s-x) > 0\\
    a & a^T(s-x) < 0 \\
    0 & \text{o.w}
    \end{cases}
\end{align}

\end{lemma}
\begin{proof}
First we argue the convexity of $f$. Observe that

\begin{align}
    f(x)
    &= \min_{u \in \cD} |a^T(u-x)|\\
    &= \min_{u \in \cD} \|u-x \|_{aa^T}.
\end{align}

The norm $\|u-x \|_{aa^T}$ is convex in both $u$ and $x$ across $\mathbb{R}^d$. So we have that $f(x)$ which is obtained by partial minimization of a convex function across a convex domain remains convex over $\mathbb{R}^d$. It follows that for any $x,y \in \mathbb{R}^d$, 

Now let $x$ be such that $\grad f(x) = 0$. Existence of such a point is guaranteed since $\cD$ in the definition of $f$ is compact.

\begin{align}
    f(y) \ge f(x) + \grad f(x)^T (y-x). \label{eq:sub}
\end{align}

We proceed to show the Lipschitzness of $f$. Let $w \in argmin_{u \in \cD} |a^T(u-x)|$. We have
\begin{align}
    f(y) - f(x)
    &= \min_{u \in \cD} |a^T (u-y)| - \min_{u \in \cD} |a^T (u-x)|\\
    &\le |a^T(w-x)| - |a^T(w-y)|\\
    &\le |a^T(x-y)|\\ \label{eq:ub}
    &\le \| a\|_2 \| x-y\|_2.
\end{align}

Since $\| a\|_2 \le \kappa$, we conclude that the function $f$ is $\kappa$ Lipschitz.

We argue that $\grad f(x) = \lamda a$ for some scalar $\lamda$. Let $b$ be a such that $a^T b = 0$. Let $z = y+\sigma b$. Notice that by the definition of $f$, we have that $f(y) = f(z)$. So,

\begin{align}
    f(z)
    &=f(y)\\
    &\ge f(x) + \grad f(x)^T (z-x)\\
    &= f(x) + \grad f(x)^T (y-x) + \sigma \grad f(x)^T b.
\end{align}
The above inequality must hold for any $\sigma$. Note that both $f(y)$ and $f(x)$ is bounded for any two points in $x,y\mathbb{R}^d$.  Further, $\grad f(x)^T (y-x)$ is also bounded due to the Lipschitzness of $f$. So if $\grad f(x)^T b$ is not zero, we can choose a $\sigma$ such that inequality is violated, leading to a contradiction in the convexity of $f$ across $\mathbb{R}^d$.
 
So $\grad f(x)^Tb=0$. This implies that $\grad f(x) = \lamda(x) a$ for some scalar $\lamda(x)$ and for any $x \in \mathbb{R}^d$.

Next, we argue that $\lamda(x) \in [-1,1]$. Combining Eq.\eqref{eq:sub} and \eqref{eq:ub} we have
\begin{align}
    |a^T(x-y)|
    &\ge \grad f(x)^T (y-x),
\end{align}
for all $x,y \in \mathbb{R}^d$. So taking $y=0$ followed by $y=2x$ leads to
\begin{align}
    |a^Tx|
    &\ge \pm \lamda(x) a^Tx.
\end{align}

Suppose $x$ is chosen such that $a^T x\neq 0$. Then the above inequality implies that $\lamda(x) \in [-1,1]$.

Let $w \in argmin_{u \in \cD} |a^T(u-x)|$. Let $s = (x+w)/2$. We have that

\begin{align}
    f(s)
    &\ge f(x) + \lamda(x) a^T(s-x). \label{eq:1}
\end{align}

Moroever,
\begin{align}
    f(s)
    &\le |a^T(w-s)|\\
    &= \frac{1}{2} |a^T(x-w)|\\
    &= f(x) - |a^T(x-s)|.\label{eq:2}
\end{align}

Combining Eq.\eqref{eq:1} and \eqref{eq:2}, we obtain
\begin{align}
    -|a^T(s-x)|
    &\ge \lamda(x) a^T(s-x).
\end{align}

Recall that when  $a^Tx \neq 0$ , $\lamda(x) \in [-1,1]$.

So we conclude that if $a^Tx \neq 0$ and $a^T(s-x) > 0$, then $\lamda(x) \le -1$. This implies that $\lamda(x) = -1$ as $\lamda(x) \in [-1,1]$ holds true.

Similarly if $a^Tx \neq 0$ and $a^T(s-x) < 0$, then $\lamda(x) \ge 1$. This implies that $\lamda(x) = 1$ as $\lamda(x) \in [-1,1]$ holds true.

Now if $a^Tx \neq  0$ and $a^T(s-x) = 0$, we can choose $\lamda(x) = 0$ as $f(z) \ge f(x) + \lamda(x) a^T (z-x) = 0$ holds true for any $z$.

If $a^Tx = 0$, $0 \in argmin_{u \in \cD} |a^T(u-x)|$ as $0 \in \cD$ is assumed to be true. So by using the previous line of arguments we conclude that $\lamda(x)=0$.
\end{proof}

\thmdelay*
\begin{proof}

 By following the arguments in \citet{Joulani2013OnlineDelay}, we have that 

\begin{align}
    \sum_{t=1}^n f_t(x_t) - f_t(u_t)
    &= \sum_{i=1}^{\tau} \sum_{k=1}^{\floor{1+\frac{n-i}{\tau}}} f_t(x_{i+(k-1)\tau}) - f_t(u_{i+(k-1)\tau}). 
\end{align}

The second summation in the above expression is the dynamic regret of instance $i$ wrt comparator sequence $\{u_{i+(k-1)\tau}\}$ with $k$ ranging from $1$ to $\floor{1+\frac{n-i}{\tau}}$. Now by triangle inequality we have that
\begin{align}
    \sum_{k=2}^{\floor{1+\frac{n-i}{\tau}}} \| u_{i+(k-1)\tau} - i+(k-2)\tau \|_1
    &\le \sum_{t=2}^n \| u_{t} - u_{t-1} \|_1 = \mathcal{TV}(u_{1:n}).
\end{align}

Thus by Theorem \ref{thm:main} we have

\begin{align}
    \sum_{t=1}^n f_t(x_t) - f_t(u_t)
    &\le \sum_{i=1}^{\tau} \tilde O(d^3(n/\tau)^{1/3} \vee 1)\\
    &\le \tilde O(d^3\tau^{2/3}n^{1/3}[\mathcal{TV}(u_{1:n})]^{2/3} \vee \tau).
\end{align}
\end{proof}

Next, we provide the version of Corollary \ref{cor:lqr} indicating the closed form expression for all the algorithm parameters.

\begin{corollary}
Let $\Sigma_\infty = U_\infty^T \Lambda_\infty U_\infty$ be the spectral decomposition of the positive semi definite (PSD) matrix $\Sigma_\infty \in \mathbb{R}^{d_u \times d_u}$. Assume the notations in Fig.\ref{fig:algo-multi}. Let the covariate matrix $A_t := [w_{t-1}^T \ldots w_{t-m}^T] \otimes \Lambda_\infty^{1/2} U_\infty$, where $\otimes$ denotes the Kronecker product. Let the bias vector $b_t:=\Lambda_\infty^{1/2} U_\infty q_{\infty;h}^*(w_{t:t+h})$. For a sequence of DAP parameters $M_{1:n}$, let $\mathcal{TV}(M_{1:n}):= \sum_{t=2}^{n}\sum_{i=1}^m \| M_t^{[i]} - M_{t-1}^{[i]}\|_1$. For a sequence of matrices $(M^{[i]})_{i=1}^m$ define $\texttt{flatten}((M^{[i]})_{i=1}^m)$ as follows:
Let $M_k^{[i]}$ be the $k^{th}$ column of $M^{[i]}$.

Let's define 
\begin{gather}
    z^k
    =
    \begin{bmatrix}
    M^k_1\\
    \vdots \\
    M^k_{d_x}
    \end{bmatrix} \in \mathbb{R}^{d_u d_x},
\end{gather}

and
\begin{gather}
    \texttt{flatten}((M^{[i]})_{i=1}^m)
    :=
    \begin{bmatrix}
    z^1\\
    \vdots\\
    z^m
    \end{bmatrix} \in \mathbb{R}^{md_u d_x}. \label{eq:MM}
\end{gather}

Let the decision set given to the ProDR.control (Fig.\ref{fig:algo-multi}) algorithm be the DAP space defined in Eq.\eqref{eq:dap}. Let $G = 2 m d_u d_x R \gamma \sqrt{d_x \wedge d_u}  \|\Lambda^{1/2} U_\infty \|_1 + 2 \frac{ \|\Lambda^{-1/2} U_\infty B^T\|_2  \| P_\infty\|_2 \sqrt{d_u} }{1-\gamma}$. Let the delay factor of ProDR.control.delayed (Fig.\ref{fig:algo-delay}) be $\tau = h$ as defined in Proposition \ref{prop:lqr}. Choose $\alpha = \sqrt{m\| \Sigma_\infty\|_{op}}$ and $L =4G^2$. Let $\tilde R$ in Theorem \ref{thm:main} be chosen as $\tilde R = R\gamma \sqrt{d_u \wedge d_x}$. Let $z_t$ be the prediction at round $t$ made by the ProDR.control.delayed algorithm. Let $M_t^{\text{alg}} := \texttt{deflatten}(z_t)$, where $\texttt{deflatten}$ is the natural inverse operation of $\texttt{flatten}$ defined above. Let $\pi := (M_1,\ldots,M_n)$ define a sequence of DAP policies. For a sequence of matrices $M$, define $\| M\|_1 := \sum_{i=1}^m \| M^{[i]}\|_1$. By playing a control $u_t^{\text{alg}}(x_t) = \pi_t^{M_t^{\text{alg}}}(x_t)$ according to Eq.\eqref{eq:policy}, we have that
\begin{align}
    R_n(M_{1:n})
    &=\sum_{t=1}^n \ell(x_t^{\text{alg}},u_t^{\text{alg}})
    - \ell(x_t^{M_{1:n}},u_t^{M_{1:n}}) = \tilde O \left(m^3 d^4 d_x^5 (d_u \wedge d_x) (n^{1/3} [\mathcal{TV}(M_{1:n})]^{2/3} \vee 1)\right),
\end{align}
where $M_{1:n}$ is a sequence of DAP policies where each $M_t \in \cM$ (eq.\eqref{eq:dap}). Further the algorithm ProDR.control.delayed also enjoys a strongly adaptive regret guarantee for any interval $[a,b] \subseteq [n]$:
\begin{align}
    \sum_{t=a}^b \ell(x_t^{\text{alg}},u_t^{\text{alg}}) - \ell(x_t^{M},u_t^{M})
    &= \tilde O((md_ud_x)^{1.5} \log n),
\end{align}
for any fixed DAP policy $M \in \cM$.

\end{corollary}
\begin{proof}
Define
\begin{align}
    X_t
    &= [w_{t-1}^T \ldots w_{t-m}^T] \otimes I_{d_u},
\end{align}
where $I_{d_u} \in \mathbb{R}^{d_u \times d_u}$ is the identity matrix and $\otimes$ denotes the Kronecker product. Clearly $X_t \in \mathbb{R}^{d_u \times m d_u d_x}$.

With these definitions, it is easy to verify that
\begin{align}
    q^M(w_{t-1})
    &= X_t z.
\end{align}

Now we return back to  losses $\hat A_t$ mentioned in Proposition \ref{prop:lqr}. Let $\Sigma_\infty = U_\infty^T \Lambda_\infty U_\infty$ be the spectral decomposition of the positive semi definite (PSD) matrix $\Sigma_\infty \in \mathbb{R}^{d_u \times d_u}$. We have that
\begin{align}
    \hat A_{t}(M;w_{t+h})
    &= \|\Lambda_\infty^{1/2} U_\infty q^M(w_{t-1}) - \Lambda_\infty^{1/2} U_\infty q_{\infty;h}^*(w_{t:t+h})\|_{2}^2\\
    &= \|\Lambda_\infty^{1/2} U_\infty X_t z - \Lambda_\infty^{1/2} U_\infty q_{\infty;h}^*(w_{t:t+h})\|_{2}^2.
\end{align}

Define
\begin{align}
    A_t 
    &:= \Lambda_\infty^{1/2} U_\infty X_t\\
    &= [w_{t-1}^T \ldots w_{t-m}^T] \otimes \Lambda_\infty^{1/2} U_\infty
\end{align}

Next, we proceed to compute a box that encloses all DAP policies of interest. We have for each $i \in [m]$,
\begin{align}
    \|z^i\|_\infty^2
    &\le \| z^i\|_2^2\\
    &= \|M^{[i]} \|_F^2\\
    &\le (d_u \wedge d_x) \| M^{[i]} \|_{op}^2\\
    &\le (d_u \wedge d_x)R^2 \gamma^2,
\end{align}
where the last line is due to the DAP policy set that we are interested in.

Thus the box $\cD_\infty(R\gamma \sqrt{d_u \wedge d_x}) := \cD_\infty(\tilde R)$ encapsulates the DAP policy space that we are interested in.

We need to compute the parameters in Theorem \ref{thm:main}. First, let's focus on computing $G$. We have for any $z_1,z_2 \in $
\begin{align}
    \|A_t(z_1+z_2) - 2b_t \|_1
    &\le 2 \|A_t \|_1 m d_u d_x\tilde R + 2 \| b_t\|_1, \label{eq:11}
\end{align}

where $b_t = \Lambda_\infty^{1/2} U_\infty q_{\infty;h}^*(w_{t:t+h})$.

We have
\begin{align}
    \| A_t \|_1
    &= \max_{i=1,\ldots,m}  \| w_{t-i}\|_\infty \|\Lambda^{1/2} U_\infty \|_1\\
    &\le \|\Lambda^{1/2} U_\infty \|_1,\label{eq:22}
\end{align}
as the disturbances obey $\| w_t\|_2 \le 1$.

We have
\begin{align}
    \| b_t \|_2
    &\le  \sum_{i=t}^{t+h} \|\Lambda^{-1/2} U_\infty B^T (A_{cl,\infty})^{i-t} P_\infty w_{i}\|_2\\
    &\le \sum_{i=t}^{t+h} \|\Lambda^{-1/2} U_\infty B^T\|_2 \|(A_{cl,\infty})^{i-t} \|_2 \| P_\infty\|_2 \|w_{i}\|_2\\
    &_{(a)}\le \|\Lambda^{-1/2} U_\infty B^T\|_2  \| P_\infty\|_2 \sum_{i=1}^h \gamma^{i-1}\\
    &\le \|\Lambda^{-1/2} U_\infty B^T\|_2  \| P_\infty\|_2 \cdot \frac{1}{1-\gamma},
\end{align}
where in line (a) we used the strong stability criterion and the fact that $\| w_t\|_2 \le 1$. Thus we have
\begin{align}
    \| b_t\|_1
    &\le \sqrt{d_u}\| b_t\|_2\\
    &\le \frac{ \|\Lambda^{-1/2} U_\infty B^T\|_2  \| P_\infty\|_2 \sqrt{d_u} }{1-\gamma}. \label{eq:3}
\end{align}

Putting together Eq.\eqref{eq:11}.\eqref{eq:22} and \eqref{eq:3} we arrive at
\begin{align}
     \|A_t(z_1+z_2) - 2b_t \|_1
     &\le 2 m d_u d_x R \gamma \sqrt{d_x \wedge d_u}  \|\Lambda^{1/2} U_\infty \|_1 + 2 \frac{ \|\Lambda^{-1/2} U_\infty B^T\|_2  \| P_\infty\|_2 \sqrt{d_u} }{1-\gamma}\\
     := G \label{eq:G}
\end{align}

Next we proceed to calculate $\alpha$ in Theorem \ref{thm:main}. Denote by $U_j$ the $j^{th}$ column of the matrix $U_\infty$. The squared norm of the $i^{th}$ row of the covariate matrix $A_t$ is given by
\begin{align}
    \sum_{k=1}^m \| w_{t-k}\|_2^2 \sum_{j=1}^{d_u} \lamda_j u_j^2[i]
    &\le \| \Sigma_\infty\|_{op} \sum_{k=1}^m \sum_{j=1}^{d_u} u_j^2[i]\\
    &= m\| \Sigma_\infty\|_{op},
\end{align}
where we used the fact the matrix $U_\infty$ is orthogonal. Thus we choose
\begin{align}
    \alpha = \sqrt{m\| \Sigma_\infty\|_{op}}.
\end{align}

By similar arguments used to reach Eq.\eqref{eq:G}, we choose
\begin{align}
    L = 4G^2
\end{align}

For a sequence of policies $M_1,\ldots,M_n$, observe that $\sum_{t=2}^{n} \|\texttt{flatten}(M_t) - \texttt{flatten}(M_{t-1}) \|_1 \le d_x \sum_{t=2}^{n} \|M_t  - M_{t-1}\|_1$. The last relation expresses the dynamic regret incurred by ProDR.control.delayed in terms of total variation of $\texttt{flatten}(M_t)$ to be bounded by total variation of the matrices themselves. 

Putting all the constants together and applying Theorem \ref{thm:delay} and Theorem \ref{thm:main} yields the Corollary.

\end{proof}

\thmlb*
\begin{proof}

Consider a system with matrices $A=0 \in \mathbb{R}^{2 \times 2}$, $B = \bigl[ \begin{smallmatrix}-1 & 0\\ 0 & -1\end{smallmatrix}\bigr]$, $R_x = \bigl[ \begin{smallmatrix}1 & 0\\ 0 & 0\end{smallmatrix}\bigr]$ and $R_u = 0 \in \mathbb{R}^{2 \times 2}$. In this setting $K_\infty = 0$ as per Eq.\eqref{eq:kinf}. We consider DAP polices (see Definition \ref{def:dap}) with $m=1$. Let the starting state be $x_1 = 0 \in \mathbb{R}^{2 \times 2}$. 

Let $y_t = \pm 1$ with probability half each. Let $w_t = [y_t,1]^T$. For a policy that chooses a control signal $u_t$ at time $t$, its next state is given by $x_{t+1} = w_t-u_t$ and $\ell_{t+1}(x_{t+1},u_{t+1}) = (u_t[1] - y_t)^2$. Hence for any algorithm, the loss is given by:
\begin{align}
    \sum_{t=1}^n \ell_t(x_t,u_t)
    &= \sum_{t=1}^{n-1} (u_t^{\text{alg}}[1] - y_t)^2. \label{eq:alg}
\end{align}

Divide the time horizon into bins of width $W$. Let the number of bins be $M := n/W$. We assume that $n/W$  is an integer for simplicity. Let the $i^{th}$ be denoted by $[s_i,e_i]$ for $i \in [M]$. Define
\begin{align}
    a_i := \frac{1}{W} \sum_{t=s_i}^{e_i} y_t.
\end{align}

We will uniformly use the same DAP policy within a bin $i$ as the comparator. This policy will be parameterized by the matrix $M_i := \bigl[ \begin{smallmatrix}0 & -a_i\\ 0 & 0\end{smallmatrix}\bigr]$

By Hoeffding's inequality and a union bound across all $M$ bins, we arrive at
\begin{align}
    a_i \in \left[-\sqrt{\frac{\log(nM/\delta)}{2W}},\sqrt{\frac{\log(nM/\delta)}{2W}} \right],
\end{align}
with probability at-least $1-\delta$. We will call this high probability event as $\cE$. Due to symmetry we have that $P(y_t = 1 | \cE) = 1/2$. So under the event $\cE$, the Bayes optimal online prediction of any algorithm as per Eq.\eqref{eq:alg} will be to set $u = [0,0]^T$. So within a bin we have that
\begin{align}
     \sum_{t=s_i}^{s_e} E[\ell_t(x_t,u_t)|\cE]
     &\ge W.
\end{align}

Now we need to upper bound the cumulative loss of the comparator within a bin. Since the policy within a bin is parameterized by $M_i$, we have that $u_t = -M_t w_{t-1} = [a_i,0]^T$ for all $t \in [s_i,e_i]$.

So we have:
\begin{align}
    E[(y_t-u_t)^2 | \cE]
    &= \frac{E[(y_t-u_t)^2] - E[(y_t-u_t)^2 | \cE^c] P(\cE^c)}{P(\cE)}\\
    &\le \frac{E[(y_t-u_t)^2]}{1-\delta},
\end{align}
where $\cE^c$ denotes complement of event $\cE$.

By bias variance decomposition, we have that
\begin{align}
    E[(y_t-u_t)^2]
    &= 1 - 1/W.
\end{align}

So the overall regret is lower bounded by
\begin{align}
    \sum_{i=1}^M\sum_{t=s_i}^{e_i} E[(y_t - u_t^{\text{alg}}[1])^2 | \cE] - E[(y_t - a_i)^2|\cE]
    &\ge \sum_{i=1}^M W(1 - \frac{1}{1-\delta}) + \frac{1}{1-\delta}\\
    &\ge M/(1-\delta) - W\delta/(1-\delta)\\
    &\ge M/2, \label{eq:lb}
\end{align}
where the last line is obtained by setting $\delta = 1/n^2$

Under the event $\cE$ with $\delta = 1/n^2$, the total variation (TV) of the sequence $a_{1:n}$ is given by:

\begin{align}
    \TV(a_{1:n})
    &\le \frac{n\sqrt{2 \log(n^4)}}{W^{3/2}}.
\end{align}

Now setting $W= \frac{n^{2/3} (8 \log n)^{1/3}}{C_n^{2/3}}$ we obtain $ \TV(a_{1:n}) \le C_n$ with probability at-least $1-1/n^2$.

For the sake of brevity let's denote $R(M_{1:n})$ (Eq.\eqref{eq:lqrRegret}) by $R_n$.

Continuing from Eq.\eqref{eq:lb}, we obtain that
\begin{align}
    E[R_n | \cE]
    &:=\sum_{i=1}^M\sum_{t=s_i}^{e_i} E[(y_t - u_t^{\text{alg}}[1])^2 | \cE] - E[(y_t - a_i)^2 | \cE]\\
    &\ge \frac{n^{1/3}C_n^{2/3}}{2(8 \log n)^{1/3}}, \label{eq:finr}
\end{align}
where the event $\cE$ occurs with probability at-least $1-1/n^2$.

Now consider the event $\cE^c$. For the purpose of obtaining a lower bound we can restrict our attention to comparators $a_{1:n}$ such that $|a_i| \le 1$ for all $i \in [n]$ and $\TV(a_{1:n}) \le C_n$. Using the DAP policy given by $M_i := \bigl[ \begin{smallmatrix}0 & -a_i\\ 0 & 0\end{smallmatrix}\bigr]$ as comparators, we have that  under the event $\cE^c$
\begin{align}
    R_n
    &\ge - \sum_{t=1}^n (y_t - a_t)^2\\
    &\ge -4n
\end{align}

So overall we have that
\begin{align}
    E[R_n]
    &\ge E[R_n | \cE] p(\cE) + E[R_n | \cE^c] p(\cE^c)\\
    &\ge  \Omega(n^{1/3}C_n^{2/3}) (1-1/n^2) - 4n \cdot (1/n^2)\\
    &= \Omega(n^{1/3}C_n^{2/3}).
\end{align}

When $C_n \le 1/\sqrt{n}$, the static regret bound of $\Omega (\log n)$ (see Theorem 11.9 in \citet{BianchiBook2006}). This completes the proof of the theorem.

\end{proof}

\textbf{Connections to online non-parametric regression framework of \citet{rakhlin2014online}.} In the work of \citet{rakhlin2014online}, they study the following online regression framework (simplified here without affecting the information-theoretic rates):

\begin{itemize}
    \item At each round $t$, learner plays a decision $x_t \in \mathbb{R}$.
    \item Nature reveals a label $y_t$ such that $|y_t| \le 1$.
    \item Learner suffers loss $(y_t-x_t)^2$.
\end{itemize}

One is interested in finding the min-max rate of regret against a non-parametric sequence class. We define the space of total variation (TV) bounded sequences as:

\begin{align}
    \TV(C_n)
    &:= \{ \theta_{1:n} | \TV(\theta_{1:n}) \le C_n \}.
\end{align}

Translated into the setup of \citet{rakhlin2014online}, one can aim to control the regret against $\TV(C_n)$ which is:

\begin{align}
    R_n
    &:= \sum_{t=1}^n (y_t - x_t)^2 - \inf_{\theta_{1:n} \in \TV(C_n)} \sum_{t=1}^n(y_t - \theta_t)^2. \label{eq:npr}
\end{align}

The TV class is known to be sandwiched between two Besov spaces having the same minimax rate (see for eg. \citep{DeVore1993ConstructiveA}). So the results of \citet{rakhlin2014online} based on characterizing the sequential Rademacher complexity of the Besov class leads to $O(n^{1/3})$ as the minimax rate of $R_n$ wrt $n$. The rate wrt $C_n$ was not provided in their work. However, we remark that they establish an $O(n^{1/3})$ upper bound also via non-constructive arguments.

In contrast, the lower bound we provided in the proof of Theorem \ref{thm:lb} is for $\sum_{t=1}^n E[(y_t - u_t^{\text{alg}}[1])^2 - (y_t - a_t)^2 | \cE]$ (Eq.\eqref{eq:finr}) where $\TV(a_{1:n}) \le C_n$ under the high probability event $\cE$ trivially lower bounds $R_n$ in Eq.\eqref{eq:npr} with high probability.

\end{document}